\documentclass[letterpaper, 10 pt, conference]{ieeeconf}  

\IEEEoverridecommandlockouts                              

\usepackage{xcolor}
\usepackage{graphicx}
\usepackage{amsmath}
\usepackage{empheq}
\usepackage{amssymb}

\usepackage{booktabs}
\usepackage{amsfonts}
\usepackage{pifont}

\usepackage{xspace}
\def\method{\textit{Diff-Muscle}\xspace}

\title{\LARGE \bf
Diff-Muscle: Efficient Learning for Musculoskeletal Robotic Table Tennis}
\author{Wentao Zhao$^{1}$, Jun Guo$^{1}$, Kangyao Huang$^{1}$, Xin Liu$^{1}$, Huaping Liu$^{1}$\footnotemark$^\text{\dag}$
\thanks{\footnotemark$^\text{\dag}$ Corresponding author: hpliu@tsinghua.edu.cn}
\thanks{$^{1}$Department of Computer Science and Technology, Tsinghua University, China.}
}

\begin{document}

\maketitle
\thispagestyle{empty}
\pagestyle{empty}

\begin{abstract}
Musculoskeletal robots provide superior advantages in flexibility and dexterity, positioning them as a promising frontier towards embodied intelligence. However, current research is largely confined to relative simple tasks, restricting the exploration of their full potential in multi-segment coordination. Furthermore, efficient learning remains a challenge, primarily due to the high-dimensional action space and inherent overactuated structures. To address these challenges, we propose \method, a musculoskeletal robot control algorithm that leverages differential flatness to reformulate policy learning from the redundant muscle-activation space into a significantly lower-dimensional joint space. Furthermore, we utilize the highly dynamic robotic table tennis task to evaluate our algorithm. Specifically, we propose a hierarchical reinforcement learning framework that integrates a Kinematics-based Muscle Actuation Controller (K-MAC) with high-level trajectory planning, enabling a musculoskeletal robot to perform dexterous and precise rallies. Experimental results demonstrate that \method significantly outperforms state-of-the-art baselines in success rates while maintaining minimal muscle activation. Notably, the proposed framework successfully enables the musculoskeletal robots to achieve continuous rallies in a challenging dual-robot setting.

\end{abstract}

\section{INTRODUCTION}
One of the primary objectives in the pursuit of generalist embodied intelligence is to develop agents capable of exhibiting human-like flexibility, dexterity, and behavioral diversity within dynamic and unstructured environments~\cite{liu2025embodied}. Compared to conventional motor-driven systems, musculoskeletal systems have recently been demonstrated to offer superior flexibility and dexterity, attributed to their inherent overactuated nature~\cite{wangmyochallenge}. However, current research on musculoskeletal robots remains largely confined to simplified tasks within isolated body segments, such as in-hand manipulation. While these scenarios provide valuable insights, they often overlook the critical importance of multi-segment coordination and fail to fully capture the extreme control dimensionality inherent in complex musculoskeletal systems. Consequently, these restricted experimental settings struggle to fully exploit the mechanical potential of musculoskeletal robots, leaving a significant gap between theoretical advantages and practical execution in complex tasks.

Robotic table tennis has long served as a premier benchmark for evaluating the integrated capabilities of robotic systems~\cite{hu2025versatilehumanoidtabletennis}, for it demands whole-body coordination, rapid reactive responses, and high-precision manipulation. Executing a successful return necessitates not only the coordination between trunk, arm, and hand but also instantaneous decision-making to intercept fast-moving balls. Such a task compels the robots to maintain a stable, active grasp while simultaneously modeling the racket's impact position, velocity, and orientation with extreme accuracy. Consequently, robotic table tennis serves as an ideal platform to develop advanced control algorithms for musculoskeletal robots, manifesting inherent potential in flexibility, dexterity, and behavioral diversity.

\begin{figure}[t]
    \centering
    \includegraphics[height=6.5cm]{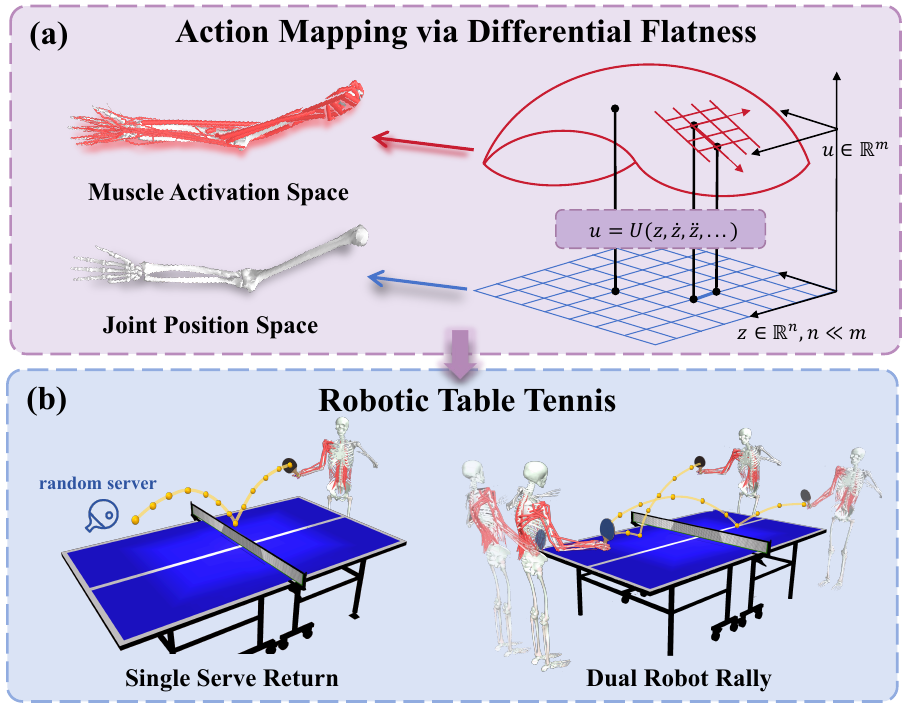}
    \caption{\textbf{Diff-Muscle.} (a) Utilizing the inherent differential flatness, \method reformulates the learning problem in musculoskeletal systems from high-dimensional muscle space to low-dimensional joint space. (b) We evaluate \method in the highly dynamic robotic table tennis task to demonstrate its capability and efficiency in learning multi-segment coordination and rapid reactive behaviors.}
    \label{fig1} 
\end{figure}

Learning control policies for musculoskeletal systems has long been a challenge due to the following reasons. First, the performance suffers from the high dimensionality and overactuated nature. For instance, the motion of the human lumbar spine is driven by over 180 individual muscles, and the hand is driven by more than 30 muscles to perform complex tasks~\cite{stokes1999quantitative, van2011constraints}. Besides, there exists a complex mapping relation between muscles and joints, where a single joint is actuated by multiple muscles, and a single muscle may simultaneously influence the states of several joints~\cite{zhang2019modeling}. Furthermore, the muscle dynamic is non-linear and characterized by its unidirectional force-generating nature, as muscles can only generate contractile forces~\cite{hill1938heat}. While these properties provide significant control flexibility, they simultaneously increase the difficulty of learning a precise control policy. 

Existing research primarily focuses on leveraging muscle synergy prior to learn a synergy representation~\cite{he2024dynsyn,berg2024sar}. Nevertheless, synergy patterns of specific muscle groups exhibit variations in different moving directions~\cite{ting2007neuromechanics}. For instance, in multi-DoF structures like the lumbar spine, a change in movement direction may shift a pair of muscles from synergy to antagonism, making it difficult to identify explicit synergy patterns from collected data. Besides, such methods fail to achieve an effective reduction in the actual dimensionality of the action space, leaving policies to explore in high-dimensional muscle space. These limitations motivate us with a question: can we leverage inherent kinematic structure and muscle dynamics in musculoskeletal systems to map redundant action spaces onto a lower-dimensional joint space, bridging the gap from joint coordination to muscle execution?

In this work, we address this challenge by reformulating the musculoskeletal control problem through the lens of differential flatness. Differential flatness~\cite{sira2018differentially} is a powerful property of certain nonlinear systems that allows the full state and input to be algebraically determined by a set of carefully chosen flat outputs and their finite derivatives. Through rigorous mathematical derivation, we prove the conditional differential flatness of musculoskeletal systems, identifying joint configurations as the ideal flat outputs. This theoretical foundation enables a principled transformation of the control manifold, effectively mapping the highly redundant and high-dimensional muscle activation space onto a significantly lower-dimensional joint space. Building upon this bridge, we propose \method (Fig.~\ref{fig1}), a framework that achieves substantial dimensionality reduction, significantly enhancing exploration efficiency while simultaneously preserving the intrinsic dynamical fidelity of musculoskeletal systems. The main contributions of this work are summarized as follows:

\begin{itemize}
    \item We propose \method, which enables direct policy learning in joint space by utilizing a Kinematics-based Muscle Actuation Controller (K-MAC), improving the exploration efficiency for musculoskeletal systems.
    \item We develop a hierarchical reinforcement learning framework for the robotic table tennis task, enabling efficient learning of complex striking behaviors.
    \item Extensive experiments demonstrate that the \method framework significantly outperforms current baselines, and we successfully enable two musculoskeletal robots to execute consecutive rallies.
\end{itemize}

\begin{figure*}[t]
    \centering
    \includegraphics[height=7.8cm]{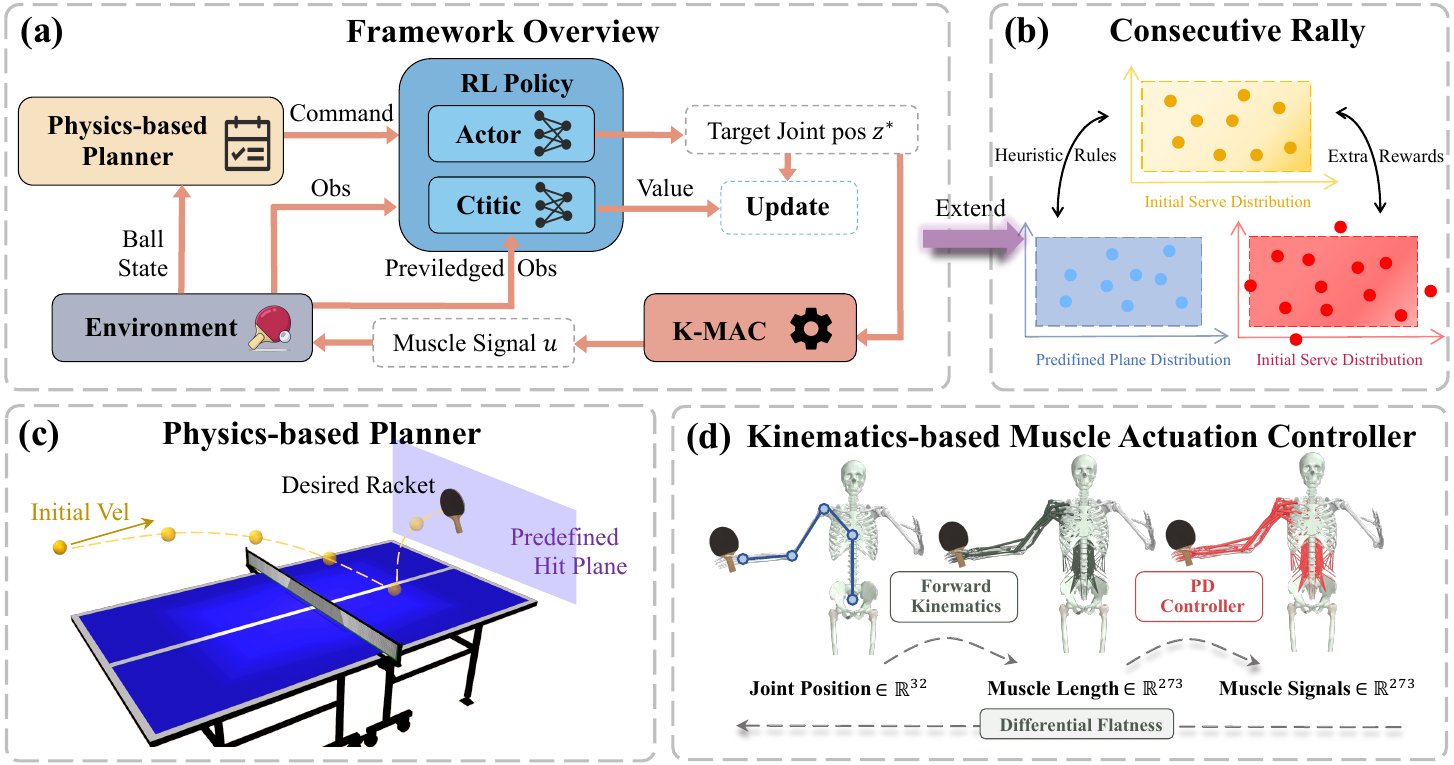}
    \caption{\textbf{Hierarchical reinforcement learning framework.} (a) In this framework, the policy takes as input current observation and high-level commands, and generates the target joint positions, which are subsequently translated into muscle signals through the Kinematics-based Muscle Actuation Controller (K-MAC). (b) We extend the framework and achieve successful consecutive dual-robot rallies. (c) Given the ball state, the planner predicts the desired racket position, orientation, and velocity at the predefined plane. (d) The Kinematics-based Muscle Actuation Controller integrates forward kinematics and a PD controller to translate target joint positions into muscle control signals.}
    \label{fig2}
\end{figure*}

\section{RELATED WORKS}

\subsection{Musculoskeletal System Control}

The control of high-dimensional, overactuated musculoskeletal systems is a longstanding challenge in both robotics and biomechanics due to the high dimensionality of action space and the redundancy of muscle actuators~\cite{zhao2025bayesian,wei2025motion}. While deep reinforcement learning has become the predominant approach for musculoskeletal systems, the vast action space often leads to poor exploration efficiency. Besides, standard reinforcement learning algorithms typically rely on independent Gaussian noise for exploration, which is suboptimal for musculoskeletal models where uncorrelated perturbations in redundant muscles might diminish each other's effects. To address this, DEP-RL~\cite{schumacher2022dep} integrates Differential Extrinsic Plasticity (DEP) to generate action correlated noise for more effective exploration. Similarly, Lattice~\cite{chiappa2023latent} injects temporally correlated noise into the latent state of the policy network to induce structured, cross-actuator correlations, effectively leveraging learned actuator synergies to improve exploration efficiency. Besides advancing exploration efficiency, some other studies leverage bio-inspired muscle synergies to mitigate the complexity of policy learning. SAR~\cite{berg2024sar} acquires synergistic action representations from behaviors learned during play and transfer these knowledge to more complex tasks. Along the same lines, Dynsyn~\cite{he2024dynsyn} generates synergistic representation directly from the system's dynamical structure via random perturbations, demonstrating superior learning efficiency across various high-dimensional musculoskeletal models.

\subsection{Robotic Table Tennis}

Robotic table tennis has long been a focal point in the robotics community, serving as an ideal testbed for evaluating control algorithms due to its inherent requirements for rapid response, agile movement, and high-precision manipulation. Early research~\cite {mahjourian2018hierarchical,ma2025mastering} primarily explored self-play algorithms, where agents iteratively refined their performance by competing against historical versions of their own policies. Recently, drawing inspiration from the hierarchical nature observed in biological motor control, a growing body of literature has leveraged hierarchical frameworks to learn more complex maneuvers in robot table tennis. Google DeepMind~\cite{ma2025mastering} proposes a hierarchical and modular policy architecture that utilizes a high-level controller to select between various low-level strike skill policies. While this approach achieves human-level performance in robot table tennis, it is validated on a relatively simple 6-DoF robot arm platform. To bridge the gap towards more complex robot systems, Hitter~\cite{su2025hitter} extends to a Unitree G1 humanoid robot, enabling consecutive strikes through hierarchical planning and learning. To move closer towards human-like dexterity and intelligence, MyoSuite~\cite{caggiano2022myosuite} proposes a challenge focusing on mastering an upper-body musculoskeletal model to rally table tennis, and our approach is validated on this platform.

\section{PROBLEM FORMULATION}

We formulate the robotic table tennis task as a Partially Observable Markov Decision Process (POMDP), defined by a tuple $\mathcal{M}=(\mathcal{S}, \mathcal{O}, \mathcal{A}, \mathcal{P}, \mathcal{R},\Omega, \gamma$), where $\mathcal{S} \in \mathbb{R}^n$ represents all valid states, $\mathcal{O} \in \mathbb{R}^m$ represents the partial observation received by the agent according to the function $\Omega$, and $\mathcal{A} \in \mathbb{R}^k$ represents continuous actions. The transition function $P(s_{t+1}|s_t, a_t)$ describes the probability of an agent taking an action $a_t$ to transfer from the current state $s_t$ to $s_{t+1}$ at time $t$. $\mathcal{R}: \mathcal{S} \times \mathcal{A} \rightarrow \mathbb{R}$ is the reward function. And $\gamma \in [0,1)$ is the discount factor. The objective is to optimize a policy $\pi_{\phi}$ that maximizes the expected discounted return:
\begin{equation}
    \pi_\phi^*(a_t|o_t) = \arg\max_\phi \mathbb{E}_{\tau \sim \pi_\phi} \left[ \sum_{t=0}^{\infty} \gamma^t r(s_t, a_t) \right]
\end{equation}
where $o_t$ is the observation at step $t$, and we optimize the policy with Proximal Policy Optimization (PPO)~\cite{schulman2017proximalpolicyoptimizationalgorithms}.

We later demonstrate (Fig.~\ref{fig:reward}) that direct end-to-end learning in the raw muscle action space fails to learn complex behaviors. This failure is attributed to the inherent challenges of high-dimensionality and the overactuated nature of musculoskeletal systems. In contrast, the differential flatness of musculoskeletal systems with respect to joint positions suggests that the learning paradigm can be algebraically simplified. In the next section, we detail our \method algorithm, which leverages this prior to learn control policy directly in lower-dimensional joint space.

\section{METHOD}

We first provide a rigorous proof of conditional differential flatness for musculoskeletal systems, establishing joint configurations as flat outputs. Building on this, we introduce the Kinematics-based Muscle Actuation Controller (K-MAC) to analytically translate joint targets into physiological activations. We then detail the hierarchical reinforcement learning framework (Fig.~\ref{fig2}) designed for robotic table tennis and finally extend this framework to achieve dual-robot rallies by decoupling striking and recovery phases.

\subsection{Differential Flatness and Action Mapping}
Controlling a musculoskeletal robot using traditional RL methods suffers from the curse of dimensionality caused by high-dimensional inputs and outputs. In this work, we leverage differential flatness to address the ultra-high-dimensional output problem, thereby reducing the output from the high-dimensional muscle activation space to a low-dimensional target joint angle space, significantly compressing the search space without information loss. 

\subsubsection{Conditional Differential Flatness}
To establish the differential flatness of the musculoskeletal system, we first define its governing dynamics and subsequently provide a constructive proof. Specially, our formulation is strictly grounded in the widely accepted Hill-type muscle model as implemented in MuJoCo~\cite{todorov2012mujoco}. The complete musculoskeletal dynamics comprises four coupled components--the skeletal dynamics, muscle-joint geometry, the Hill-type muscle model, and muscle activation dynamics:
\begin{equation}
    M(z)\ddot{z} + C(z,\dot{z})\dot{z} + G(z) = \tau_m + \tau_{ext},
    \label{eq:dynamic}
\end{equation}
\begin{equation}
    \tau_m = J_m^\top(z) f_{m},
    \label{eq:tau_m}
\end{equation}
\begin{equation}
    f_{m} = act \cdot F_{max} \cdot F_L(l_{m}) \cdot F_V(\dot{l}_{m}) + F_P(l_{m}),
    \label{eq:f_mi}
\end{equation}
\begin{equation}
    \frac{\partial act(t)}{\partial t} = \frac{u(t) - act(t)}{\tau_{u,act}}
    \label{eq:activation}
\end{equation}
where $z \in \mathbb{R}^n$ denotes the joint angles and $\tau_{ext}$ is the external torque; $\tau_m$ denotes the resultant torque generated by muscles; $f_m \in \mathbb{R}^m$ is the muscle tension vector; $F_{max}$ is the peak active force at zero velocity; $act \in [0,1]^m$ and $u \in [0,1]^m$ are muscle activation state and the bounded control input respectively; $J_m(z)=\partial \ell_{FK}(z)/ \partial z$ is the transposition of the muscle Jacobian matrix; $F_L$ denotes the active force-length function, $F_V$ denotes force-velocity function, and $F_P$ denotes passive force function. Besides, $\tau_{u,act}$ is the time constant computed at runtime as:
\begin{equation}
    \tau_{u, act} = 
    \begin{cases} 
    \tau_{act} \cdot (0.5 + 1.5 \cdot act), & u - act > 0 \\ 
    \tau_{deact} / (0.5 + 1.5 \cdot act), & u - act \leq 0 
    \end{cases}
\end{equation}
where $\tau_{act}$ and $\tau_{deact}$ are with defaults $(0.01,0.04)$.

\textit{Theorem 1} \textbf{Conditional Differential Flatness}. Considering Eq.\ref{eq:dynamic}--Eq.\ref{eq:activation}, the system is differentially flat with flat output $y=z$ if and only if: (C1) \textit{Kinematic solvability}: $\mathrm{rank}(J^\top_m(z))=n, z \in \mathcal{Z}$; (C2) \textit{Physiological feasibility}: $act \in [0,1]^m$; (C3) \textit{Velocity non-singularity}: $F_V(\dot l_{m})\ne 0$ for all active muscles.

\textit{Proof.} (Sufficiency) Let the flat output be $y=z$. The state variables are trivially expressed as $z=y$ and $\dot z=\dot y$. From Eq.\ref{eq:dynamic}, the required joint torque is obtained as
\begin{equation}
    \tau_m = M(y)\ddot{y} + C(y,\dot{y})\dot{y} + G(y) - \tau_{ext} =: \mathcal{T}(y,\dot{y},\ddot{y},\tau_{ext})
\end{equation}
where the symbol $=:$ indicates ``is defined as". The muscle kinematics follow from the forward kinematics mapping $\ell_{FK}$:
\begin{equation}
    l_m = \ell_{FK}(y), \quad \dot{l}_m = J_m(y)\dot{y}.
\end{equation}
Consequently, the force-length, force-velocity, and passive force functions can be expressed as:
\begin{align}
    F_L(l_{m}) &= F_L(\ell_{FK}(y)) =: \mathcal{F}_{L}(y),\\
    F_V(\dot{l}_{m}) &= F_V(J_m(y)\dot{y}) =: \mathcal{F}_{V}(y,\dot{y}),\\
    F_P(l_{m}) &= F_P(\ell_{FK}(y)) =: \mathcal{F}_{P}(y).
\end{align}
By condition (C1), $J^\top_m$ is full row rank, hence the linear system Eq.\ref{eq:tau_m} admits the general solution:
\begin{equation}
    f_m = (J_m^\top)^+ \tau_m+P_\mathcal{N}\cdot \alpha
\end{equation}
where $(\cdot)^+$ denotes the Moore-Penrose pseudoinverse, and $P_\mathcal{N}=I-(J_m^\top)^+J^\top_m$ is the projection onto the null space $\mathcal{N}(J^\top_m)$ which includes all combinations of muscle forces that do not generate joint torque, and $\alpha \in \mathbb{R}^m$ is a free parameter. Substituting $\tau_m=\mathcal{T}(y,\dot{y},\ddot{y},\tau_{ext})$ yields 
\begin{equation}
    f_m = \mathcal{F}(y,\dot{y},\ddot{y},\tau_{ext}).
\end{equation}
Finally, solving Eq.\ref{eq:f_mi} for the activation input under condition (C3) gives
\begin{equation}
    act = \frac{f_{m} - \mathcal{F}_{P}(y)}{F_{max}\mathcal{F}_{L}(y)\mathcal{F}_{V}(y,\dot{y})} =: \mathcal{A}(y,\dot{y},\ddot{y},\tau_{ext}).
\end{equation}
Condition (C2) ensures $\mathcal{A} \in [0,1]$. Furthermore, the system control input $u$ gives
\begin{align}
    u = &
    \begin{cases} 
    act + \dot{act} \cdot \tau_{act} \cdot (0.5 + 1.5 \cdot act), & u - act > 0 \\
    act + \dot{act} \cdot \tau_{deact} / (0.5 + 1.5 \cdot act), & u - act \leq 0 
    \end{cases} \\ 
     &\Rightarrow u=: \mathcal{U}(y, \dot{y}, \ddot{y}, \dddot{y}, \tau_{ext})
\end{align}
Thus, all states of the system and $u$ are algebraically determined by $y$ and its derivatives up to third order, establishing differential flatness for Eq.\ref{eq:dynamic}--Eq.\ref{eq:activation}. This completes the proof.

\textit{Proof.} (Necessity) If (C1) fails, Eq.\ref{eq:tau_m} has no solution. If (C3) fails, $act$ cannot be algebraically determined from Eq.\ref{eq:f_mi}. If (C2) fails, the feasible activation set is empty, and the muscle cannot be activated. Thus, all three conditions are necessary.

\subsubsection{Kinematics-based Action Mapping}
According to the conditional differential flatness of the system, we propose a kinematics-based muscle actuation controller, which integrates forward kinematics to translate target joint configurations into physiological muscle activations. As illustrated in Fig.~\ref{fig2}(d), given the target joint positions $z^*$ from $\pi_{\phi}$, the desired muscle lengths $l_m^*$ can be derived through the forward kinematics of the musculoskeletal system:
\begin{equation}
    l^*_m = \ell_{FK}(z^*).
\end{equation}
To compute the muscle forces required to achieve these desired muscle lengths, we then employ a PD controller to generate the required muscle force based on current muscle length $l_m$ and velocity $\dot{l}_m$:
\begin{equation}
  f_{m} =   F_{max} \left( K_p (l_{m}^* - l_{m}) - K_d \dot l_{m} \right)/\Delta l_{m}
\end{equation}
where $K_p$ is proportional gain parameter and $K_d$ is derivative parameter, and both are hyperparameters. $\Delta l_{m}$ is the length range. Subsequently, the muscle force is clipped within a physiologically plausible range:
\begin{equation}
 f_{m}^* =  \text{CLIP}\left( f_{m},\; -F_{max},\; 0 \right)
\end{equation}
Given the desired muscle force $f_{m}^*$, the corresponding muscle activation $act$ and system input $u$ can be computed by Eq.\ref{eq:f_mi} and Eq.\ref{eq:activation} respectively.

\subsection{Physics-based Planner}

\subsubsection{Planner Construction}
Generating actions directly from a fast-moving ball presents significant challenges, as RL policies require extensive undirected exploration to develop an understanding of ball trajectories and learn to return the ball at appropriate positions. However, acquiring the future states of the ball allows the policy to proactively plan current movements and advance the reward signals. Drawing inspiration from Hitter~\cite{su2025hitter}, we implement a Physics-based Planner (Fig.~\ref{fig2}(c)) to predict the incoming ball's trajectory and compute the target racket position, velocity, and orientation as high-level commands for the policy $\pi_{\phi}$.

Given the initial ball position $\mathbf{p}$ and velocity $\mathbf{v}$, we model its flight and table-collision dynamics, ignoring air resistance and spin:

\begin{subequations} \label{eq:ball_dynamics}
\begin{empheq}[left={\empheqlbrace}]{align}
&\dot{\mathbf{v}} = \mathbf{g}, &  &\text{if } p_z > 0 \label{eq:flight} \\
&\mathbf{v}^+ = \mathbf{C} \mathbf{v}^-, &  &\text{if } p_z = 0 \label{eq:bounce}
\end{empheq}
\end{subequations}
where $\mathbf{g}$ is gravitational acceleration, $\mathbf{v}^+$ and $\mathbf{v}^-$ represent the velocity before and after impact with the table, and $\mathbf{C} = \mathrm{diag}(C_h, C_h, -C_v)$ is the restitution matrix with horizontal $C_h$ and vertical $C_v$ coefficients. The planner integrates Eq.~\eqref{eq:ball_dynamics} forward until the ball intersects a predefined virtual hitting plane, yielding the predicted striking position $\hat{\mathbf{p}}_\text{strike}$.

Assuming the striking velocity is normal to the racket plane, we sample the target return velocity $\mathbf{v}_o$. We first sample the vertical component $v_z$ randomly from a predefined range, and derive the flight duration $t$ via the flight dynamics. Subsequently, we determine the feasible velocity bounds in the $xy$-plane to ensure the ball lands within the opponent's table area while crossing the net:

\begin{equation}
v_{x} \sim \mathcal{U}(\frac{x_{lower}}{t}, \frac{x_{upper}}{t})
\end{equation}

\begin{equation}
v_{y} \sim \mathcal{U}(\frac{y_{lower}}{t}, \frac{y_{upper}}{t})
\end{equation}

Assuming frictionless contact with a restitution coefficient $C_r$ along the normal direction of the racket surface, the desired racket velocity at contact is computed as:

\begin{equation}
\hat{\mathbf{v}}_\text{racket} = \frac{\mathbf{v}_o \cdot \mathbf{u} + C_r\, \mathbf{v}_i \cdot \mathbf{u}}{1 + C_r} \, \mathbf{u},
\quad
\mathbf{u} = \frac{\mathbf{v}_o - \mathbf{v}_i}{\|\mathbf{v}_o - \mathbf{v}_i\|}.
\end{equation}

With $\hat{\mathbf{v}}_\text{racket}$, we can compute the corresponding quaternion $\hat{\mathbf{o}}_\text{racket}$. The resulting tuple $(\hat{\mathbf{p}}_\text{racket}, \hat{\mathbf{v}}_\text{racket}, \hat{\mathbf{o}}_\text{racket})$ constitutes the high-level command sent to the policy as observation.
\subsubsection{Reward Functions}

Based on the high-level commands, our reward function is a weighted sum of three components:
\begin{equation}
r = w_gr_g + w_cr_c + w_sr_s
\end{equation}
where $w_g, w_c, w_s $ are the corresponding reward weights. A dense grasp reward $r_g$ is used to minimize the distance between the fingers and the handle to prevent dropping during rapid movements. The command reward $r_c$ provides dense feedback for the target racket pose tracking and a sparse velocity term for precise hitting speeds. Finally, we introduced an additional sparse reward $r_s$ related to task completion. A sparse reward is granted when the ball makes contact with the racket. Immediately after bounce, we utilize the planner to predict the landing position on the other side, assigning a sparse reward if a successful return is predicted.

\subsection{Extend to Dual Robot Rally}

Building upon the successful implementation of single-serve returns, we further extend our framework to a continuous dual-robot rally task to better approximate realistic, multi-round table tennis scenarios. A critical challenge in this transition is the compounding errors over multi-exchanges which causes the ball's trajectory to gradually drift outside the initial distribution, eventually making it unreachable and terminating the rally.

To address this challenge, we must ensure the policy consistently faces in-distribution incoming balls. As illustrated in Fig. \ref{fig2}(b), we constrain the ball's trajectory distribution within identical spatial boundaries across distinct planes. Specifically, we first employ a heuristic serving method to restrict the ball distribution on the predefined striking plane within the serving range. Moreover, an additional sparse reward is triggered only when the returned ball reaches the opponent's striking plane within identical spatial boundaries. The extended framework prevents trajectory drift over multiple rounds and effectively supports stable, long-horizon exchanges.

\section{EXPERIMENTS}

\subsection{Experimental Setup and Task Formulation}

\subsubsection{Musculoskeletal Model}
\begin{figure}[t]
    \centering
    \includegraphics[height=5.7cm]{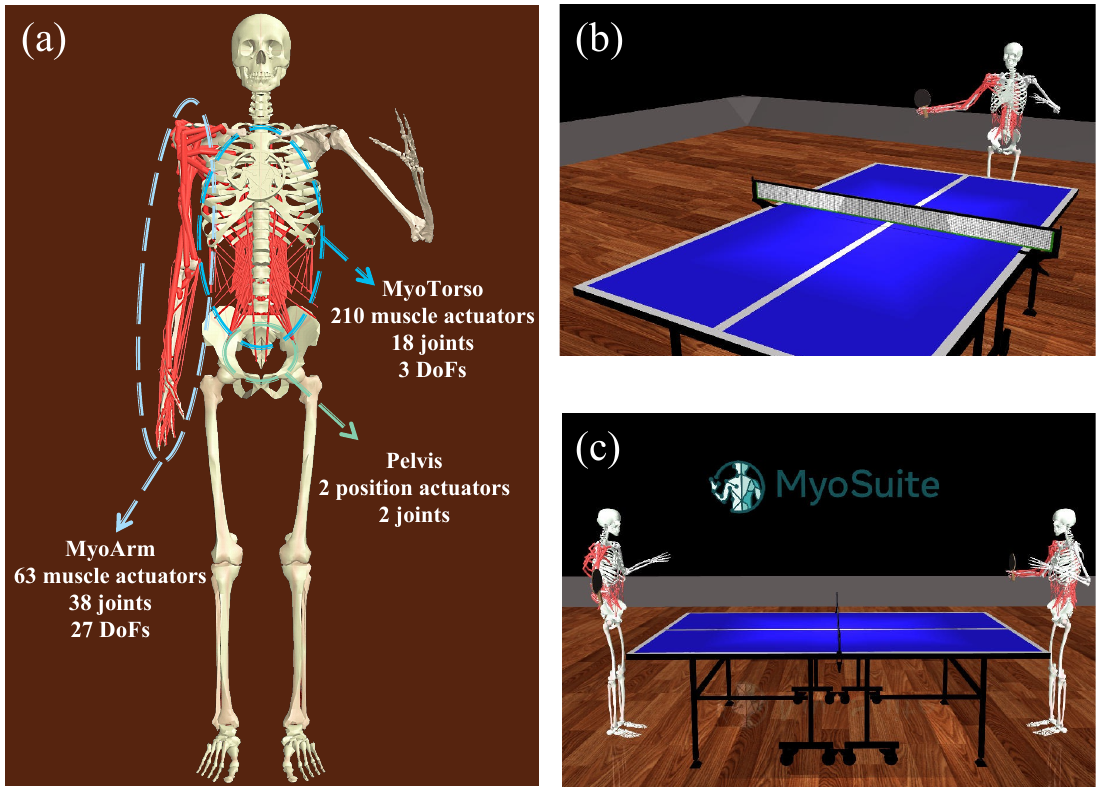}
    \caption{\textbf{Musculoskeletal Model and Environment.} (a) The musculoskeletal model consists of MyoArm (63 muscles, 27 DoF), MyoTorso (210 muscles, 3 DoF), and pelvis (2 actuators, 2DoF). (b) Single-robot environment to execute a single-serve ball return. (c) dual-robot rally environment. }
    \label{fig:model}
\end{figure}

We utilize the upper-body musculoskeletal model from the \textit{MyoSuite} environment \cite{caggiano2022myosuite} to validate our proposed framework. As illustrated in Fig.~\ref{fig:model}(a), the robot is a highly complex, overactuated system comprising a total of \textbf{273 muscle actuators} that control \textbf{58 joints}. The model is primarily composed of the following components:

\begin{itemize}
    \item \textbf{MyoArm}: The arm consists of 63 muscles and 38 joints. With 11 dependent joints passively driven by the active ones, it provides 27 degrees of freedom (DoFs).
    \item \textbf{MyoTorso}: The torso contains 210 muscles and 18 joints. Under equality constraints, it effectively operates with only \textbf{3 DoFs}, representing a severely overactuated system where coordinating high-dimensional muscle redundancy is exceptionally challenging.
    \item \textbf{Pelvis}: The pelvis is equipped with two position actuators, allowing the robot to translate across the horizontal plane ($x, y$ directions).
\end{itemize}

\subsubsection{Task Definition and Environments}
As shown in Fig.~\ref{fig:model}(b), our single-robot experimental environment is built upon the \textit{MyoChallenge 2025} table tennis track, where a musculoskeletal robot is required to execute high-speed table tennis returns. The task is formulated as a single-serve ball return problem with the following specifications:

\begin{itemize}
    \item \textbf{Environment Initialization}: At the beginning of each episode, a ball is initialized on the opponent's side within a predefined spatial range. Its initial velocity is sampled to ensure a valid trajectory that lands on the agent's side of the table. The agent is reset to a consistent initial posture, with the paddle placed at the grasping position of the hand. Crucially, the paddle is not rigidly attached to the hand, necessitating active grasping through muscle coordination.
    \item \textbf{Success Criteria}: Each episode consists of a single serve. A trial is considered successful if and only if the ball is hit by the paddle exactly once and subsequently lands within the boundaries of the opponent's court.
    \item \textbf{Control Objectives}: The task demands coordination between the \textit{MyoArm} and \textit{MyoTorso} models. The agent must apply appropriate muscle forces to maintain a stable grasp of the paddle while executing high-speed, accurate strikes to return the ball to the opponent's area.
\end{itemize}

Furthermore, to evaluate our framework in a setting that more closely reflects real-world table tennis play, we build a dynamic dual-robot rally environment by extending the single-robot setup (Fig.~\ref{fig:model}(c)). Specifically, we replace the static ball server with a symmetric agent. Each episode begins with a randomized serve from either side, transitioning into a continuous rally that terminates upon a missed hit or a rule violation. This setup shifts the task from a static distribution to a non-stationary distribution, where each incoming ball is inherently conditioned on the previous agent's action. Consequently, the agent must not only master precision striking but also maintain stability after striking to be ready for the next rally.

\subsection{Effectiveness of Kinematics-based Muscle Actuation Controller}
This section evaluates the effectiveness of the proposed Kinematics-based Muscle Actuation Controller in mitigating the learning challenges inherent in high-dimensional musculoskeletal systems. To ensure a fair comparison across baselines, all experiments are conducted in the single-robot table tennis environment using consistent reward parameters and hyperparameters. All results are averaged across five random seeds to ensure statistical significance.

\subsubsection{Baseline Selection and Evaluation Metrics}
We compare our method against several state-of-the-art reinforcement learning methods for musculoskeletal control: 
\begin{itemize}
    \item \textbf{Muscle-PPO}: A standard PPO algorithm where the policy directly outputs activation values for all 273 muscles and 2 pelvis actions.
    \item \textbf{Muscle-PD~\cite{feng2023musclevae}}: Inspired by MuscleVAE, this policy outputs target muscle lengths, which are subsequently mapped to muscle forces via a PD controller and finally converted into muscle activations.
    \item \textbf{DynSyn~\cite{he2024dynsyn}}: This algorithm derives muscle-grouping representations from system dynamics via random perturbations and uses state-dependent weights for coordinated control.
\end{itemize}
Our simulation is built on mjlab~\cite{zakka2026mjlablightweightframeworkgpuaccelerated}, which provides a simulation wrapper around GPU-accelerated physics simulation MuJoCo Warp with CUDA Graph optimization. All algorithms adopt PPO as the basic algorithm.

We conduct 1000 independent test episodes for each trained policy. To evaluate the experiment performance, we utilize three primary metrics: \textbf{hit rate} is defined as the ratio of successful contact between ball and racket over all independent test episodes, while \textbf{success rate} corresponds to the ratio of valid return to the opponent's side. And \textbf{effort} is used to quantify control efficiency. We define effort as the average normalized muscle activations across all evaluated episodes: 
\begin{equation}
\mathrm{Effort} = \frac{1}{N} \sum_{e=1}^{N} \left( \frac{1}{T_e} \sum_{t=1}^{T_e} \frac{\lVert \mathbf{a}_{e,t} \rVert_2}{n_a} \right)
\end{equation}
where $\mathbf{a}_{e,t}$ represents the muscle activation vector at time step $t$ of episode $e$, $n_a$ denotes the number of actuators, $T_e$ is the duration of episode $e$, and $N$ is the total number of episodes. Lower effort values indicate more efficient and energy-conscious control strategies.

\subsubsection{Learning Efficiency in High-dimensional System}
\begin{figure}[t]
    \centering
    \includegraphics[height=4.5cm]{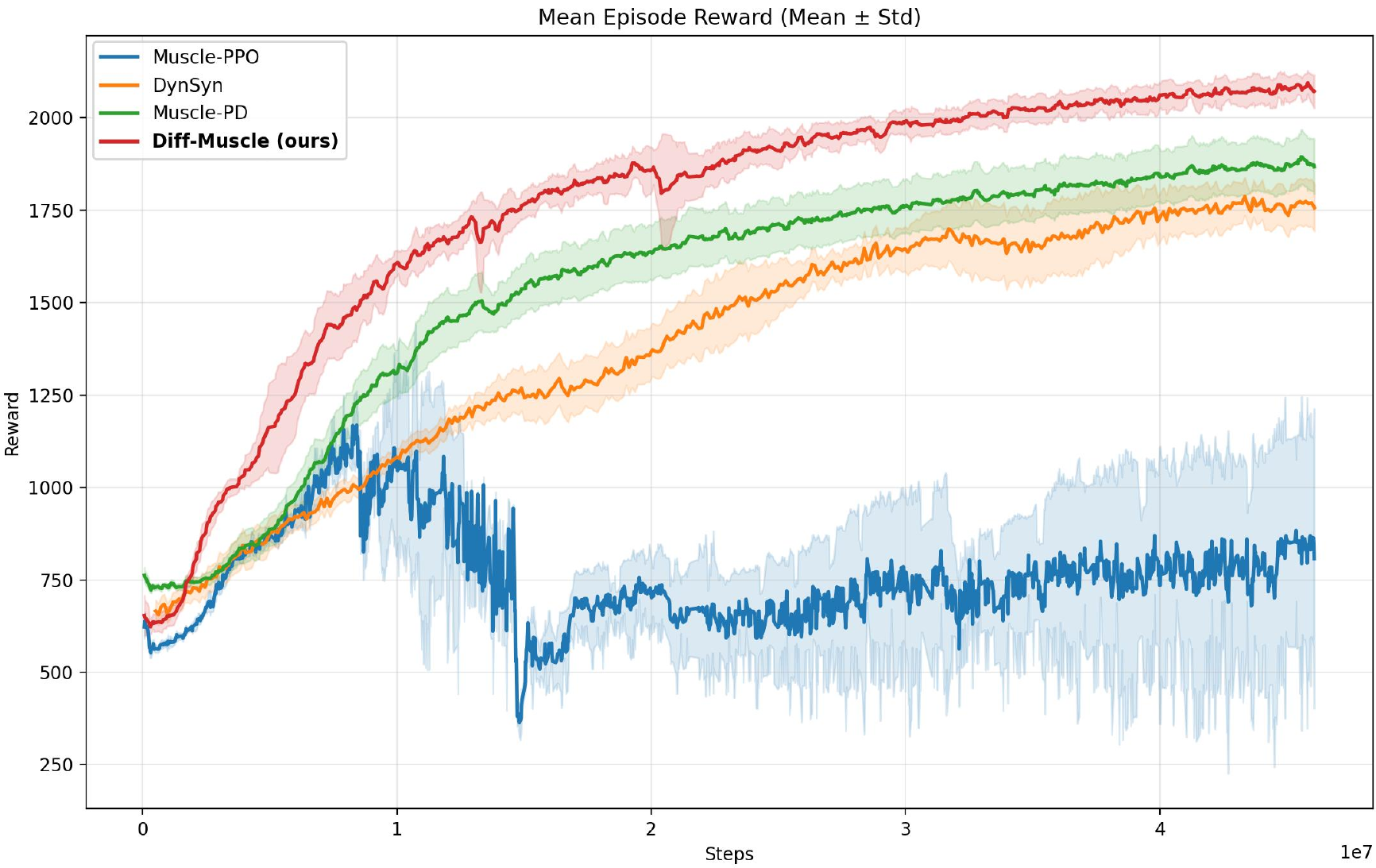}
    \caption{\textbf{Learning curve.} \method exhibits the most rapid improvement in reward and achieves the highest overall performance among all evaluated methods. Mean $\pm$ Std is computed across 5 random seeds. }
    \label{fig:reward}
\end{figure}

Fig.~\ref{fig:reward} illustrates the training progress of \method and baselines in terms of mean episode reward curves. As depicted, \method exhibits the most rapid rise and achieves the highest converged reward of approximately 2100. These results demonstrate that \method possesses significantly higher exploration efficiency, enabling both efficient and stable policy learning. In contrast, the baselines struggle to achieve comparable performance due to the complexities of exploring in a high-dimensional space. Notably, the end-to-end Muscle-PPO baseline fails to discover effective action patterns, suffering from a catastrophic performance collapse after its initial progress. Additionally, the Dynsyn baseline suffers from a slower learning efficiency and inferior final reward, suggesting that learned synergies are less effective than our kinematic priors for such a dynamic task.

\begin{table}[t]
\setlength\tabcolsep{3pt}
\linespread{1.2}
\begin{center}\small
\caption{Comparison with existing methods on the single-robot environment.}

\begin{tabular}{c|ccc}
\hline\hline
Method & Hit(\%) & Success(\%) & Effort \\
\hline\hline
{Muscle-PPO} & 76.4 & 43.0 & 0.035 \\
{DynSyn~\cite{he2024dynsyn}} & 84.5 & 62.6 & 0.0310 \\
{Muscle-PD~\cite{feng2023musclevae}} & 90.5 & 72.9 & 0.0365 \\
{\textbf{Diff-Muscle (ours)}} & \textbf{95.1} & \textbf{80.3} & \textbf{0.0278}\\

\hline\hline
\end{tabular}
\label{experiment1}
\end{center} 
\end{table}

\subsubsection{Task Performance and Bio-mechanical Efficiency}
Table~\ref{experiment1} presents the evaluation results of our \method and baselines. As illustrated, \method achieves the highest hit rate of 95.1\% and success rate of 80.3\%. From the test results, we find that \method exhibits superior stability and adaptability, consistently adopting appropriate striking postures to return difficult balls, such as some high-altitude balls. Even in extreme scenarios where a successful return is nearly impossible, the agent still prioritizes making contact with the racket. While other baselines fail when facing high-velocity serves or balls aimed at extreme positions, such as those aimed at the far backhand side. 

Furthermore, the third column of Table~\ref{experiment1} demonstrates that \method realizes the lowest muscle activation energy. This bio-mechanical efficiency is primarily achieved through our kinematics-based muscle actuation controller by setting a relatively low proportional gain $k_p$, while the other baselines exhibit substantially higher muscle activation levels, manifesting their inefficiency in terms of energy consumption.

\begin{figure}[t]
    \centering
    \includegraphics[height=4.2cm]{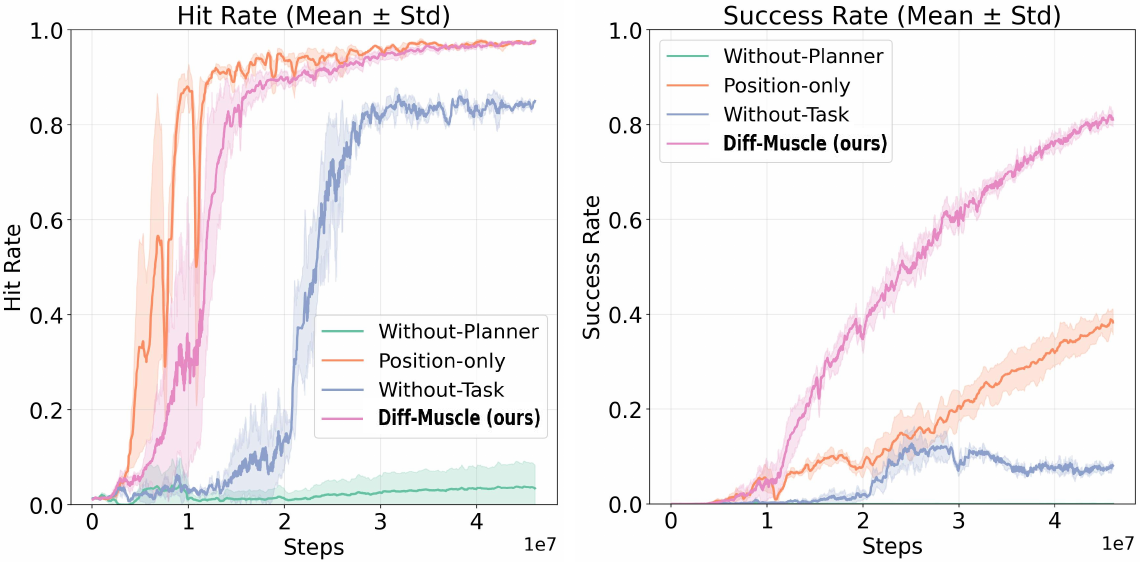}
    \caption{\textbf{Ablation results.} \method exhibits the most rapid improvement in reward and achieves the highest overall performance among all evaluated methods. Mean $\pm$ Std is computed across 5 random seeds. }
    \label{reward2}
\end{figure}

\subsection{Impact of Predictive Planning}
This section aims to validate the effectiveness of predictive planning of our framework in the table tennis task by ablating several modules. The experiments are similarly conducted in the single-robot environment, with results averaged across five random seeds.

\subsubsection{Variants Settings}

We designed three variants to individually evaluate the contributions of different modules:
\begin{itemize}
    \item \textbf{Without-Planner}: This variant drops the physics-based planner entirely, excluding its predicted commands from both the observation and the reward function.
    \item \textbf{Position-only}: This variant solely utilizes the striking position $\hat{\mathbf{v}}_\text{racket}$ predicted by the physics-based planner as the component for observations and reward function.
    \item \textbf{Without-Task}: This variant excludes the sparse task reward $r_{s}$ associated with successful task completion.
\end{itemize}

\begin{figure}[t]
    \centering
    \includegraphics[height=4.2cm]{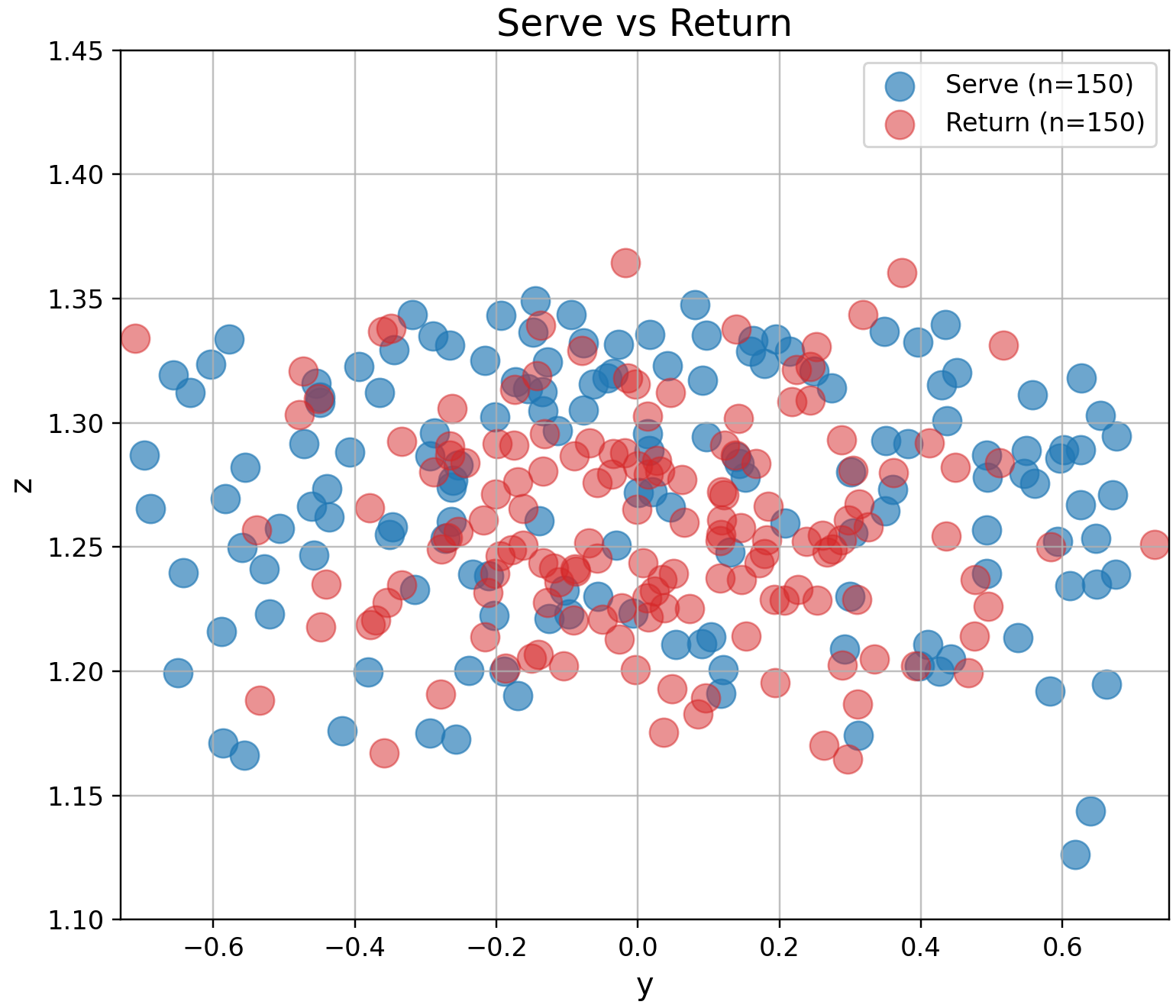}
    \caption{\textbf{Distribution visualization.} This scatter plot visualizes the y-z coordinates of 150 serves and corresponding returns. }
    \label{compare_dist}
\end{figure}

\begin{figure*}[t]
    \centering
    \includegraphics[height=4.3cm]{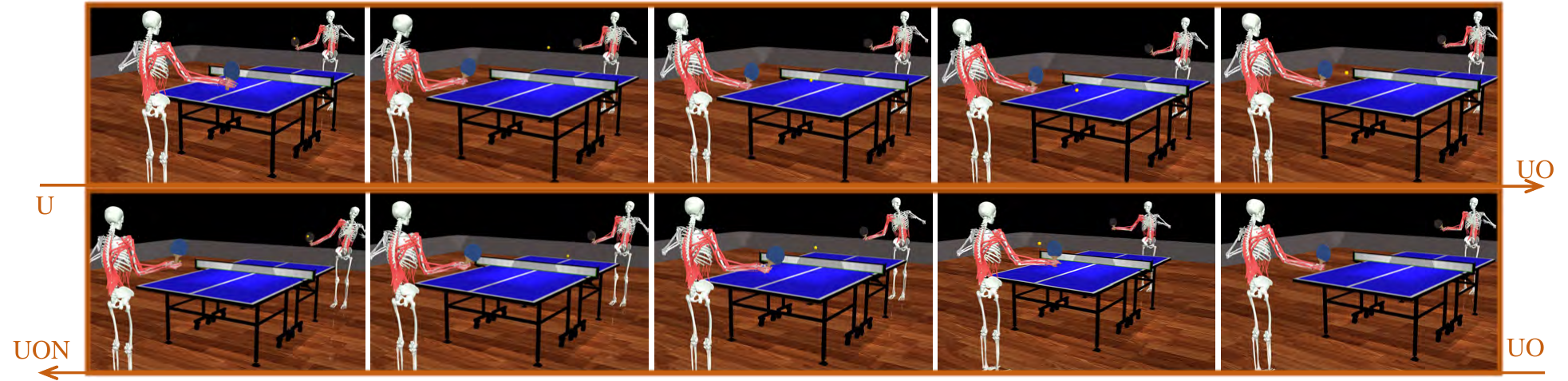}
    \caption{\textbf{Visualization of dual-robot rally.} In a round of continuous rally, the robot actively moves to the target hitting position and hits the ball at a certain speed, returning it to a position where the opponent can receive it.   }
    \label{dual}
\end{figure*}

\subsubsection{Necessity of Predictive Planning on Reactive Tasks}
As illustrated in Fig.~\ref{reward2}, the Without-Planner variant completely fails to learn the task, with both its hit rate and success rate remaining near zero. Without trajectory prediction, the policy relies entirely on instantaneous observations, resulting in delayed reactive behaviors that struggle to intercept fast-moving balls in time.

Furthermore, the Position-only variant presents a compelling phenomenon. While it learns to intercept the ball exceptionally fast, achieving a nearly 100\% hit rate early in training, its success rate severely plateaus around 40\%. This indicates that lacking the proactive preparation of target racket velocity and orientation makes it difficult for the agent to explore the precise return trajectory. These results underscore that predicting the complete kinematic command is essential to bridge the gap between mere interception and a valid, successful return.

\subsubsection{Importance of Task-oriented Reward}
The Without-Task variant highlights the indispensability of the sparse task-related reward $r_s$. As shown in Fig.~\ref{reward2}, relying solely on the dense command tracking reward $r_c$ enables the agent to achieve a relatively high hit rate of approximately 85\%, but its success rate is severely degraded, hovering merely around 10\%. This proves that while tracking commands effectively guide the racket to the predicted spatial target, they are insufficient for mastering the complex contact dynamics. The sparse task reward provides critical feedback for fine-tuning the impact velocity and micro adjustments in orientation when contacting the ball. This fine-tuning is what ultimately transforms a simple collision into a successful return.

\subsection{Dual-Robot Rally}
To further evaluate our extended framework's performance in the consecutive dual robot rally task, we conduct additional experiments in the dual-robot environment. 

Firstly, to investigate the stability of the dual-robot rally, we analyzed the spatial trajectories of the balls as it returns the initial serve plane. As depicted in Fig.~\ref{compare_dist}, the distribution of the returned balls (red) closely aligns with the initial serve distribution (blue) across 150 evaluated episodes. This overlap indicates that our policy consistently encounters in-distribution balls, effectively preventing the "trajectory drift" that leads to early termination in multi-exchange scenarios.

A representative sequence of these consecutive rallies is visualized in Fig.~\ref{dual}. In each round, the robot proactively coordinates its torso and arm to move the racket to the target hitting position and strike the ball with an appropriate speed, ensuring it returns to a position where the opponent can successfully receive it. However, despite the alignment of ball trajectories, failures still occur during extended rallies. We observed that while our policy effectively maintains the ball within the 'in-distribution' spatial range, it does not guarantee a $100\%$ success rate for every individual strike, which terminates the rally. Finally, experimental results demonstrate that our approach empowers two musculoskeletal robots to successfully achieve up to 14 consecutive exchanges, through the maintenance of long-term trajectory consistency, making a meaningful step toward achieving athletic intelligence in complex robotic systems.

\section{CONCLUSION}

We presented \method, an algorithm that enables efficient learning for musculoskeletal systems. Through the conditional differential flatness of such systems, we reformulate the learning problem from high-dimensional muscle space to joint space. Besides, we integrate \method and develop a hierarchical reinforcement learning framework for musculoskeletal robotic table tennis. Experimental results demonstrate that our framework outperforms current baselines and achieves up to 14 consecutive rallies in a challenging dual-robot setting, marking a substantial step toward bio-inspired athletic intelligence.

\bibliographystyle{IEEEtran}
\bibliography{IEEEexample}

\end{document}